# Representation learning with CGAN for causal inference

Zhaotian Weng[1*], Jianbo Hong[2], Lan Wang[3]

[1]School of Software, Tsinghua University, Beijing China,
wengzt18@mails.tsinghua.edu.cn

[2]Collage of Computer Science and Technology, Jilin University, Changchun China,
hongjb2419@mails.jlu.edu.cn

[3]College of Engineering, Monash University, Suzhou China, dolores4048@gmail.com

*Correspondence author email: wengzt18@mails.tsinghua.edu.cn

**Abstract**— Conditional Generative Adversarial Nets (CGAN) is often used to improve conditional image generation performance. However, there is little research on Representation learning with CGAN for causal inference. This paper proposes a new method for finding representation learning functions by adopting the adversarial idea. We apply the pattern of CGAN and theoretically demonstrate the feasibility of finding a suitable representation function in the context of two distributions being balanced. The theoretical result shows that when two distributions are balanced, the ideal representation function can be found and thus can be used to further research.

## 1.    Introduction

The causal inference has many significant applications in real life. Can a drug work for a specific patient population? Does the promulgation of a policy have an impact on leadership approval ratings? Will online classes have an impact on student's academic performance? These are all questions of causal inference [1]. It is difficult to answer the above questions only by relying on data, and we can only answer the questions of relevance by relying on data. Causal inference refers to a task considering the assumptions, estimation strategies, and study designs that are useful to draw conclusions based on data. Causal inference depends on the consideration of counterfactual states. Mainly, it considers the outcomes that could manifest given exposure to each set of treatment conditions [2]. The process of causal inference always contains two steps: the representation step and the prediction step. The representation step is to make the distribution between the factual distribution and counterfactual distribution similar, and we will explain the reason in the next chapter. The prediction step is to predict the potential outcome based on the input x and condition t. This paper further introduces a conditional GAN [3] model to the traditional causal inference representation learning model to obtain a better representation function to balance the data distribution, improving the prediction step following. The paper is organized as follows: Chapter 2 introduces our work, Chapter 3 and Chapter 4 provide information about Generative Adversarial Network and Conditional Generative Adversarial Network, respectively, Chapter 5 illustrates the advantage of our work and Chapter 6 gives the conclusions and the future work.

## 2. Our work

### 2.1. Purpose

The main problem faced in causal inference is the imbalance of the data distribution. We make the following definition: $X$ is the set of contexts, T is the set of possible actions, $Y$ is the set of possible outcomes, and for t, which can be either 0 or 1, $Y_t(x)$ belonging to $Y$ is the potential outcome for x belonging to $X$. The quantity of interest is Individual Treatment Effect defined as:

$$ITE(x) = Y_1(x) - Y_0(x).$$

The essential problem of causal inference is we can only observe $Y_t(x)$ for one specific value of $t$, so we have to predict the potential outcome $Y_t(x)$ for the other $t$. Moreover, the main problem is that we train on a dataset with a factual, empirical distribution but predict on a counterfactual empirical distribution; the distributions of these two datasets are often dissimilar [4]. Therefore, it is necessary to introduce a representation learning function to characterize the input data to make their data distribution similar. In this post, we hope to optimize representation learning methods to improve the performance of causal inference models.

### 2.2. Adversarial Nets

Previous work used neural networks to select representation functions; we added the idea of CGAN [5] to the model, GAN is a new way to train generative models. The new model consists of two adversarial models, which are a generative model $G$ and a discriminative model $D$; both $G$ and $D$ can be nonlinear mapping functions.

Generative adversarial nets are composed of two adversarial models one of which is a generative model, and the other one is a discriminative model. The generative model named $G$ aims to capture the data distribution and the discriminative model known as $D$ aims to estimate the probability that a sample came from the training data instead of $G$. The generator constructs a mapping function from a noise distribution $p(z)$ to data space as $G(z; \Theta_g)$. Also, a single scalar representing the probability that $x$ came from data was given by the discriminator $D$ named $D(x; \Theta_g)$. Finally, the G and D would be trained following the formula shown below[6]:

$$\min_G \max_D V(D, G) = E_{x \sim p_{data}(x)}[\log D(x)] + E_{z \sim p_z(z)}[\log(1 - D(G(z)))]$$

## 3. Conditional Adversarial Nets

The generative adversarial nets could be expanded to conditional adversarial nets with some extra information known as conditions which could be labels or any other formats were given to both the generator and discriminator.

The noise $p(z)$ and extra information $y$ were combined in the joint hidden representation, the adversarial training framework was highly flexible in how the representation was composed [6].

The $x$ and $y$ would be represented as inputs to function $D$. $D$ and $G$ would play the two-player min-max game with the value function as following:

$$\min_G \max_D V(D, G) = E_{x \sim p_{data}(x)}[\log D(x|y)] + E_{z \sim p_z(z)}[\log(1 - D(G(z|y)))]$$

As far as we know, we are the first to use CGAN to optimize the representation function to get better results of causal inference.

## 4. Adversarial Nets For Representations

In this part, we first introduce the value function of GAN, then, in order to find our wanted representation learning function, we apply the CGAN [7] idea and perform conditional adversarial nets for representations. Concretely, we use the noise and the control group data to find the representation function for treatment group data. Finally, the procedure of the training is displayed in Algorithm 1.

## 5. Generative Adversarial Nets

Generative Adversarial Network (GAN) includes Generator ($G$) and Discriminator ($D$) two parts. The former maps random noise to samples while the latter discriminates generated and actual samples. On the one hand, the generator $G$ strives to capture the data distribution and generate samples that are hard to differentiate by discriminator $D$. Its purpose is to maximize the probability that $D$ will make a mistake. On the other hand, the discriminator $D$ is optimized to identify real data from generated ones. The training procedure resembles a two-player versus game with the following value function,

$$\min_G \max_D F(D,G) = E_{t \sim p_{data}(t)}[\log D(t)] + E_{w \sim p_w(w)}[\log(1 - D(G(w)))] \quad (1)$$

In which $t$ is a ground truth sample from the true distribution $p_{data}$, $w$ is a noise variable sampled over the distribution $p_w$.

## 6. Adversarial Nets for Representations

As we mentioned before, our objective is to find a representation function to characterize the input data to balance the data distribution of treatment and control groups. Therefore, we borrow the idea of CGAN and try to use adversarial nets to find our wanted representation function $\Phi(t)$. In the context of the classic causal inference question, we have a treatment group and a control group of two existing anti-diabetic medications, A or B is better for a given patient. Our work aims to use the CGAN idea to find a representation learning function that makes the data distribution of the treatment group similar to the control group.

First, we define $\Phi(t)$ as the representation learning function, $t$ is the input treatment group data. To get the distribution $p_g$ of the generator over $\Phi(t)$, we set a prior on input control group data $p_{Con}(Con)$, and $G(Con; \theta_g)$ represents the mapping to data space. The perceptron $\theta_g$ with multiple layers represents the different function G. Also, we set another multi-layer perceptron $D(\Phi(t); \theta_d)$. Then, we denote $z$ as the noise variable, which will be fed into both the discriminator and generator units as an additional input layer to enhance the robustness. In the generator, control group data $p_{Con}(Con)$, and $z$ are incorporated in the shared hidden representation. In the discriminator, $\Phi(t)$ and $w$ are presented as inputs and to a discriminative function. The objective function of the two-player versus game would be represented as Eq2:

$$\min_G \max_D F(D,G) = E_{\Phi(t) \sim p_{data}(\Phi(t))}[\log D(\Phi(t)|w)] + E_{Con \sim p_{Con}(Con)}[\log(1 - D(G(Con|w)))] \quad (2)$$

During the training, we turn the last term of (2) into $-E_{Con \sim p_{Con}(Con)}[\log(D(G(Con|w)))]$ to enhance the stability of CGAN.

We use an iterative and numerical approach to implement the game. In particular, we perform N steps to optimize D and one step to optimize G in a training iteration. While the G changing slowly enough, the result in D will be maintained near its best solution. The procedure of the training formally displays in Algorithm 1.

---

**Algorithm 1** Minibatch stochastic gradient descent training of conditional adversarial nets for representations. The number of steps applied to the discriminator, N, is a hyperparameter.

___________________________________________________

**for** number of training iterations **do**

    **for** N steps **do**

        ● Sample minibatch of $s$ examples $\{Con^{(1)}, \ldots, Con^{(s)}\}$ from $p_g(Con)$.

        ● Sample minibatch of $s$ examples $\{\Phi(t)^{(1)}, \ldots, \Phi(t)^{(s)}\}$ from treatment group which produces distribution $p_{data}(\Phi(t))$.

        ● Sample minibatch of $s$ noise samples $\{w^{(1)}, \ldots, w^{(s)}\}$ from noise distribution.

        ● Ascend the stochastic gradient to update the discriminator:

$$\nabla_{\theta_d} \frac{1}{s}\sum_{i=1}^{s} [\log D(\Phi(t)^{(i)}|w^{(i)}) + \log (1 - D(G(Con^{(i)}|w^{(i)})))] \qquad \text{end for}$$

- Sample minibatch of examples $\{Con^{(1)}, \ldots, Con^{(s)}\}$ from $p_g(Con)$.
- Sample minibatch of noise samples $\{w^{(1)}, \ldots, w^{(s)}\}$ from noise distribution.
- Descend the stochastic gradient to update the generator :

$$\nabla_{\theta_g} \frac{1}{s}\sum_{i=1}^{s} \log (1 - D(G(Con^{(i)}|w^{(i)})))$$

**end for**

## 7. Theoretical Results

This section will present the theoretical result of our work, i.e., applying CGAN to the finding of a representation learning function that treatment group data distribution is similar to control group data distribution. As the samples $G(Con)$ acquired when $Con \sim p_{Con}$, the generator G sets a distribution $p_g$. We will show that this min-max game has the global optimum in the context of $p_g = p_{data}$, and Algorithm 1 can obtain the desired result by optimizing Eq.2. It should be aware that this section's setting is non-parametric.

Firstly, we intend to determine the optimum discriminator D given any generator G.

**Proposition 1.** *For any given G, the optimal discriminator D is*

$$D_G^*(\Phi(t)) = \frac{p_{data}(\Phi(t))}{p_{data}(\Phi(t)) + p_g(\Phi(t))} \qquad (3)$$

*Proof.* The discriminator D intends to maximize the value function $F(G,D)$.

$$F(G,D) = \int_{\Phi(t)} p_{data}(\Phi(t)) \log(D(\Phi(t)|w)) d\Phi(t) + \int_{Con} p_{Con}(Con) \log(1 - D(g(Con|w))) dCon$$

$$= \int_{\Phi(t)} p_{data}(\Phi(t)) \log(D(\Phi(t)|w)) + p_g(\Phi(t)) \log(1 - D(\Phi(t)|w)) d\Phi(t)$$

(4)

For any $(m,n) \in \mathfrak{R}^2 \setminus \{0,0\}$ function $f \to m\log(f) + n\log(1-f)$ reaches its maximum in $[0,1]$ at $\frac{m}{m+n}$. The discriminator only needs to be set inside of $Supp(p_{data}) \cup Supp(p_g)$, and the proof is concluded.

Thus, the Eq.2 can be formulated as:

$$H(G) = \max_D F(G,D) = E_{\Phi(t) \sim p_{data}}[\log D_G^*(\Phi(t))] + E_{Con \sim p_{Con}}[\log(1 - D_G^*(G(Con)))] =$$

$$E_{\Phi(t) \sim p_{data}}[\log D_G^*(\Phi(t))] + E_{\Phi(t) \sim p_g}[\log(1 - D_G^*(\Phi(t)))] = E_{\Phi(t) \sim p_{data}}\left[\log \frac{p_{data}(\Phi(t))}{p_{data}(\Phi(t)) + p_g(\Phi(t))}\right] +$$

$$E_{\Phi(t) \sim p_g}\left[\log \frac{p_g(\Phi(t))}{p_{data}(\Phi(t)) + p_g(\Phi(t))}\right] \quad (5)$$

**Theorem 1.** *The global minimum of H(G) is reached if and only if $p_g = p_{data}$, i.e., the data distribution of control group equals to that of treatment group. At that point, $H(G) = -\log 4$.*

Proof. From Eq.5 we observe that

$$H(G) = E_{\Phi(t) \sim p_{data}}\left[\log \frac{p_{data}(\Phi(t))}{p_{data}(\Phi(t)) + p_g(\Phi(t))}\right] + E_{\Phi(t) \sim p_g}\left[\log \frac{p_g(\Phi(t))}{p_{data}(\Phi(t)) + p_g(\Phi(t))}\right] =$$

$$E_{\Phi(t) \sim p_{data}}\left[\log \frac{p_{data}(\Phi(t))}{(p_{data}(\Phi(t)) + p_g(\Phi(t)))/2} - \log 2\right] + E_{\Phi(t) \sim p_g}\left[\log \frac{p_g(\Phi(t))}{(p_{data}(\Phi(t)) + p_g(\Phi(t)))/2} - \log 2\right] =$$

$$KL(p_{data} || \frac{p_g + p_{data}}{2}) + KL(p_g || \frac{p_g + p_{data}}{2}) - \log 4$$

(6)

In which KL is the Kullback-Leibler divergence [8]. We obtain the Jensen-Shannon divergence [9] between the distribution of the treatment group representation and the control group data distribution:

$$H(G) = -\log(4) + 2 \cdot JSD(p_{data} \| p_g) \quad (7)$$

For the reason that the Jensen-Shannon divergence between two distributions is always zero and non-negative only if they are equal, we can arrive at the conclusion that $H^* = -\log(4)$ is the global minimum of $H(G)$, and only under the circumstance where $p_{data} = p_g$ can achieve that, i.e., the only solution for the situation that the control group data distribution and distribution of treatment group representation function are balanced.

Besides, consider Eq.3, for $p_{data} = p_g$, $D_G^*(\Phi(t)) = \frac{1}{2}$. As Eq.5 suggests, $H(G) = -\log(4)$, which is the minimum of H$(G)$.

**Proposition 2.** *If G and D have enough capacity, the discriminator D can achieve its optimum given G at every step of Algorithm 1, and $p_g$ is updated so that the improvement can be made to the criterion $E_{\Phi(t) \sim p_{data}}[\log D_G^*(\Phi(t))] + E_{\Phi(t) \sim p_g}[\log(1 - D_G^*(\Phi(t)))]$ then $p_g$ converges to $p_{data}$*

As a result, in the context of $p_{data} = p_g$, i.e., the control group data distribution and distribution of treatment group representation function are balanced, we will find the most suitable representation function $\Phi(t)$ for the treatment group data, which can make control group data distribution and treatment group data distribution balanced.

## 8. Advantages & Disadvantages

Markov chains are not needed anymore which is one of our advantages and also inference is not necessary during the process of learning, the gradients is obtained only by backpropagation.[7]

Comparing to using a value such as disc to measuring the imbalance between treatment group and control group, our method is a good way to balance distributions to improve the performance of causal inference models, it can reflect the effect of representation more perfectly.

The disadvantages are that the representation model with GAN is hard to optimize. And the neural network for representation learning is not good enough and needed improving. The nerual network for prediction is also far from satisfying which can not fully reflect the actual information gain in real world, and this will do no good to the whole model.

## 9. Conclusions & Future work

Future work
1. Adjust the structure of representation neural network and prediction neural network to improve the performance of causal inference.
2. Distribution balance is not equal to the individual-level balance, we expect more viable solutions of tackling the balance on individual-level.
3. Direction left for future work is making CGAN more robust to noise.[10]

In this paper, we propose the value function V(G,D) in terms of representation to tackle the problem that data distributions in causal inference are not balanced. We also show the feasibility of using CGAN pattern to find our wanted representation learning function, i.e., a representation function in the situation where the representation distribution equals the other data distribution, and the value is -log4 in this situation. In the further research, we will explore the actual representation function by conducting experiments.

**Acknowledgement**
Zhaotian Weng, Jianbo Hong and Lan Wang contributed equally to this work and should be considered co-first authors.